\documentclass{article}
\usepackage{spconf,amsmath,graphicx}
\usepackage{bm}
\usepackage{amssymb}
\usepackage{cite}

\title{Gated Multi-layer Convolutional Feature Extraction Network for Robust Pedestrian Detection}
\name{Tianrui Liu, Jun-Jie Huang, Tianhong Dai, Guangyu Ren and Tania Stathaki\thanks{This work was supported by the EU H2020 TERPSICHORE project ``Transforming Intangible Folkloric Performing Arts into Tangible Choreographic Digital Objects'' under the grant agreement 691218.}}
\address{Department of Electrical and Electronic Engineering \\
Imperial College London, United Kingdom}
\begin{document}
%
\maketitle
\begin{abstract}
Pedestrian detection methods have been significantly improved with the development of deep convolutional neural networks. Nevertheless, robustly detecting pedestrians with a large variant on sizes and with occlusions remains a challenging problem. 
In this paper, we propose a gated multi-layer convolutional feature extraction method which can adaptively generate discriminative features for candidate pedestrian regions. 
The proposed gated feature extraction framework consists of squeeze units, gate units and a concatenation layer which perform feature dimension squeezing, feature elements manipulation and convolutional features combination from multiple CNN layers, respectively. 
We proposed two different gate models which can manipulate the regional feature maps in a channel-wise selection manner and a spatial-wise selection manner, respectively. 
Experiments on the challenging CityPersons dataset demonstrate the effectiveness of the proposed method, especially on detecting those small-size and occluded pedestrians.

\end{abstract}
\section{Introduction}
\label{sec:introduction}

Pedestrian detection has long been an attractive topic in computer vision with significant impact on both research and industry. Pedestrians detection is essential for scene understanding, and has a wide applications such as video surveillance, robotics automation and intelligence driving assistance systems. 
The pedestrian detection task is often challenged by a pedestrians with large variation of poses, appearances, sizes and under real life scenarios with complex backgrounds. 



Traditional pedestrian detectors \cite{HOG, ACF_2014dollar2014, whatcanhelpPed, HowFar-2016, DPM, Tianrui-DSP2016-DPM}
exploit various hand-engineered feature representations, such as Haar \cite{Haar-ped1997}, local binary pattern \cite{LBP-2002PAMI} as well as the Histogram of Oriented Gradient (HOG) feature and its variations. These feature representations are used in conjunction with a classifier, for instance support vector machine \cite{SVM} and boosted forests \cite{ExtremeRandomTree2006}, to perform pedestrian detection via classification. 
Recent advances of deep neural networks have made significant improvements on pedestrian detection methods \cite{IsfasterRCNNPed,2018scale-awareRCNN,CityPersons, Tianrui-DSP2017_semantic, myBMVC2018}. 
Zhang \textit{et al.} \cite{CityPersons} tailored the well known Faster-RCNN \cite{fasterRCNN2015} object detector in terms of anchors and feature strides to accommodate for pedestrian detection problems. Multi-layer Channel Features (MCF) \cite{MCF_TIP2017} and RPN+BF \cite{IsfasterRCNNPed} proposed to concatenate feature representations from multiple layers of a Convolutional Neural Network (CNN) and replace the downstream classifier of Faster R-CNN with boosted classifies to improve the performance on hard sample detection. Compared to the traditional methods, CNN-based methods are equipped with more powerful feature representation. The challenge of pedestrian detection regrading pose and appearance variations can be addressed well in most circumstances. While there is still a lot of room for improvement fro detecting pedestrian under large variations in scale. 

The visual appearance and the feature representation of large-size and small-size pedestrians are significantly different. For this reason, it is intuitive to use different feature representation for detecting objects of different sizes. In \cite{myBMVC2018}, it has been claimed that the features that can best balance feature abstraction level and resolution are from different convolutional layers. A Scale-Aware Multi-resolution (SAM) CNN-method \cite{myBMVC2018} was proposed which achieves good feature representation by choosing the most suitable feature combination for pedestrians of different sizes from multiple convolutional layers. The limitation of this method is that 
how the multi-layer feature is combined is hand-designed. Hence, there has only be a limited number of heuristic and fixed feature combinations. 

In this paper, we aim at investigating a more advanced approach which can automatically select combinations of multi-layer features for detecting pedestrians of various sizes. 
A pedestrian proposal network is used to generate pedestrian candidates and thereafter we propose a gated feature extraction network which can adaptively provide discriminative features for the pedestrian candidates of different size. 
In the proposed gated multi-layer feature extraction framework, a squeeze unit is applied to reduce the dimension of Region of Interests (RoI) feature maps pooled from each convolutional layer. It is an essential component in the gated feature extraction network in order to achieve a good balance on performance and computational and memory complexity. Following, a gate unit is applied to decide whether features from a convolutional layer is essential for the representation of this RoI. 
We investigate two gate models to manipulate and select features from multiple layers, namely, a spatial-wise gate model and a channel-wise gate model. 
We expect that the features manipulated by the proposed two gate models will be able to have stronger inter-dependencies among channels or among spatial locations, respectively. 
Experimental results show that the proposed method achieves the state-of-the-art performance on the challenging CityPersons dataset \cite{CityPersons}.



\medskip

\noindent

\begin{figure}[t]
\centering 
\includegraphics[width=0.93\columnwidth]{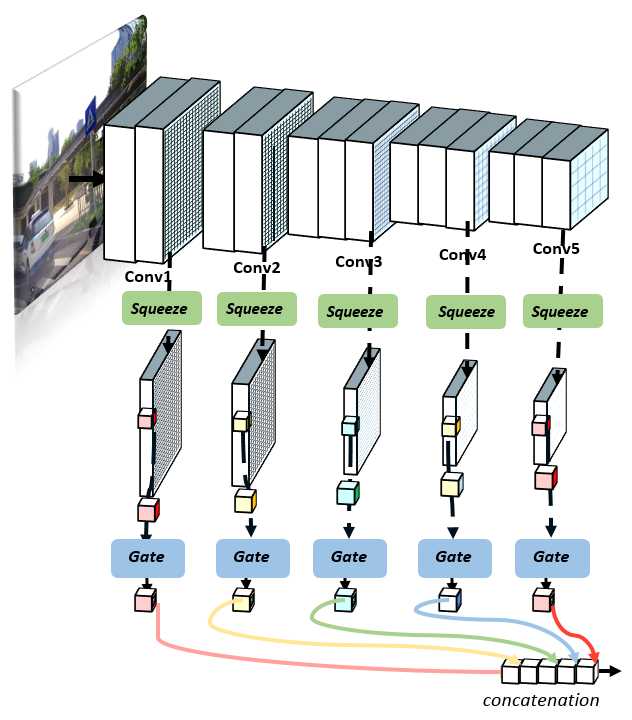}
\caption{Overview of the proposed gated multi-layer convolutional feature extraction method. Feature maps from each convolutional block of the backbone network are first passed through a \textbf{squeeze unit} for dimension reduction. The squeezed feature maps are then passed through \textbf{gate units} for feature manipulation, and finally integrated using a concatenation layer. 
}
\label{fig:gated_squeeze_multilayer_fusing} 
\end{figure}

\section{Proposed Method}


\subsection{Overview of the Proposed Detection Framework}

The proposed gated multi-layer feature extraction network aims to generate discriminative features for pedestrians with a wide range of scale variations by end-to-end learning. The overview of the proposed method is shown in Fig.\ref{fig:gated_squeeze_multilayer_fusing}. 
An input image, it is first passed through the backbone network for convolutional feature extraction.  We employ the VGG16 network \cite{VGG16} as the {backbone network}. There are 13 convolutional layers in VGG16 which can  be  regarded  as  five  convolutional blocks, i.e., $Conv1$, $Conv2$, $Conv3$, $Conv4$, and $Conv5$. Features from different layers of a CNN represent different levels of abstraction and meanwhile has different reception fields which can provide different cues for pedestrian detection. Our gated network takes the features from all the five convolutional layers as the input and will thereafter select the most discriminative feature component for pedestrian candidate of different size.
A Region Proposal Network (RPN) is used to generate a number of candidate pedestrian proposals. Given the candidate proposals, the gated multi-layer feature extraction network manipulates the CNN feature maps from each convolutional block and generates representative features for each region of interest (RoI).

The proposed gated multi-layer feature extraction network helps to realize an automatic re-weighting of multi-layer convolutional features.
Nevertheless, the gated network requires additional convolutional layers which induce a deeper RoI-wise sub-network at the cost of higher complexity and higher memory occupation. To remedy this issue, our gated sub-network includes a squeeze unit which reduces the dimension of the feature maps.
As illustrated in Fig. \ref{fig:gated_squeeze_multilayer_fusing}, features maps from each convolutional block of the backbone network are first compressed by a \textbf{squeeze unit}, then the RoI features pooled from the squeezed lightweight feature maps are passed through \textbf{gate units} for feature selection, and finally integrated at the concatenation layer. 

\subsection{The Squeeze Unit} 


A squeeze unit is used to reduce the input feature dimension of the RoI-wise sub-network in the proposed gated feature extraction network.  
Let us denote the input feature maps as  $\bm{F} = \left[{\bm{f}}_{1}, \ldots, {\bm{f}}_{C_{in}}\right] \in \mathbb{R}^{H \times W \times C_{\text{{in}}}}$
which has spatial size $H \times W$ and is of $C_{\text{{in}}}$ channels. The squeeze unit will map the input feature maps $\bm{F} \in \mathbb{R}^{H \times W \times C_{\text{{in}}}}$ 
 to the lightweight output feature maps 
$\widehat{\bm{F}} = \left[\widehat{\bm{f}}_{1}, \ldots, \widehat{\bm{f}}_{C_{\text{{out}}}}\right] \in \mathbb{R}^{H \times W \times C_{\text{{out}}}}$ with $C_{\text{{out}}} < C_{\text{{in}}}$
by applying 1 by 1 convolution, i.e.,
\begin{equation}
    \widehat{\bm{f}}_i = \bm{v}_i * \bm{F},
\end{equation}
where $\bm{v}_i$ is the $i$-th learned filter in the squeeze network for $i=1,\cdots,C_{\text{{out}}}$, and `$*$' denotes convolution.





The squeeze ratio is defined as  $r = {C_{\text{{in}}}}/{C_{\text{{out}}}}$.
In Section \ref{sec:sqeeze_ratio}, we will show that a properly selected squeeze ratio $r$ will reduce the RoI-wise sub-network parameters without noticeable performance deduction.


RoI-pooling \cite{fasterRCNN2015} will be performed on the squeezed lightweight feature maps. The features then pass through a gate unit for feature selection. 
The gate units manipulates the CNN features to highlight the most suitable feature channels or feature components for a particular RoI, while suppressing the redundant or unimportant ones. 

\subsection{The Gate Unit} 
\label{sec:gate_unit}


A gate unit will be used to manipulate RoI features pooled from the squeezed lightweight convolutional feature maps.
Generally, a gate unit consists of a convolutional layer, two fully connected (fc) layers and a Sigmoid function at the end for output normalization. 
Given regional feature maps $\bm{R}$, the output of a gate unit $\bm{G} \in \mathbb{R}^{h_g \times w_g \times c_g}$ can be expressed as:
\begin{equation}
    \bm{G}=\sigma\left(\bm{W}_{3} \delta\left(\bm{W}_{2} \delta\left(\bm{W}_{1} * \bm{R}\right)\right)\right),
\end{equation}
where $\sigma(\cdot)$ denotes the Sigmoid function, $\delta(\cdot)$ denotes the ReLU activation function  \cite{ReLU2010}, and $\bm{W} = \{ \bm{W}_1, \bm{W}_2, \bm{W}_3 \}$ are the learnable parameters of the gate network. 

The output of a gate unit $\bm{G}$ is used to manipulate the regional feature maps $\bm{R}$ through an element-wise product:
\begin{equation}
    \bm{\widehat{R}}= \bm{G} \otimes \bm{{R}},
\end{equation}
where $\otimes$ denotes the element-wise product. 

The manipulated features outputs from a gate network $\bm{\widehat{R}}$  will has the same size as its input RoI feature $\bm{R}$, and will enhance the information that is helpful for identifying the pedestrian within this RoI.

We have designed two type gate units based on how the RoI feature maps will be manipulated, namely, the spatial-wise selection gate model and the channel-wise selection gate model. 
The channel-wise selection gate model are able to increase the inter-dependencies among different features channels, while the spatial-wise selection gate model enhance the feature capacity in terms of spatial locations.

\begin{figure}[t]
    \centering
    \includegraphics[height=0.32\columnwidth]{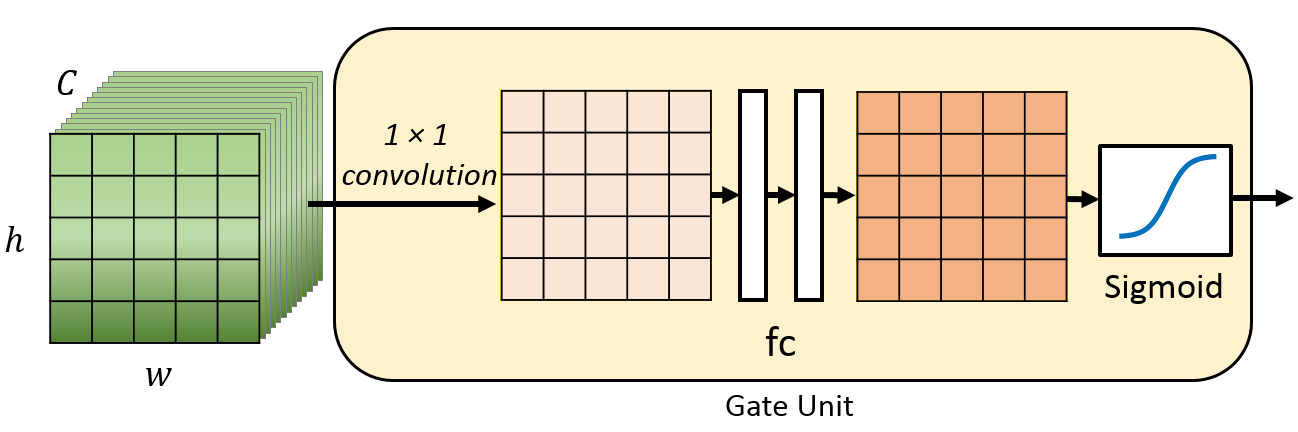}
    \caption{The spatial-wise selection gate unit. The squeezed RoI-pooled features will be transformed to a 2D map via a 1 by 1 convolution. The 2D map is followed with two fully connected layers to generate another 2D map to be fed into the Sigmoid function. 
    }
    \label{fig:gate_spatial}
\end{figure}


\begin{figure}[t]
    \centering
    \includegraphics[height=0.32\columnwidth]{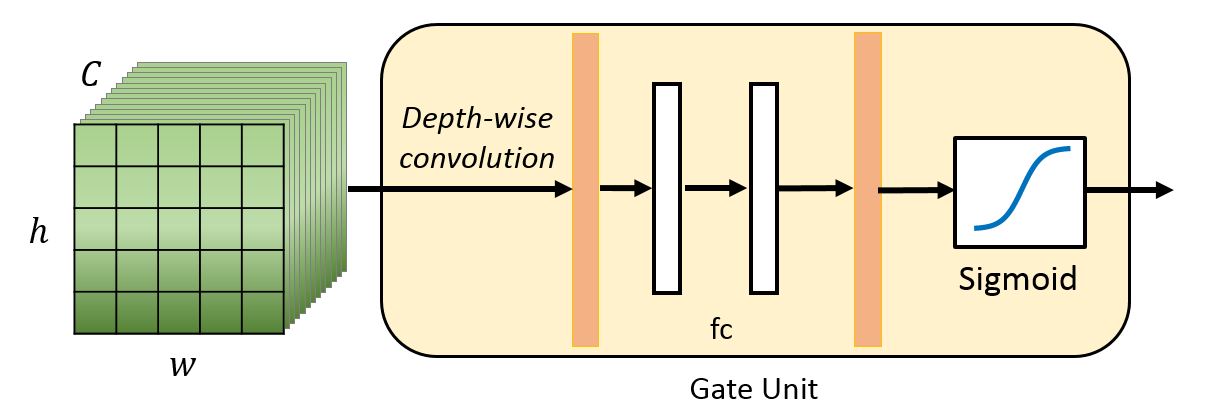}
    \caption{The channel-wise selection gate unit. The squeezed RoI-pooled features will be transformed to a 1D vector via depth-wise separable convolution. This vector is followed with two fully connected layers to generate another vector to be fed into the Sigmoid function. 
    }
    \label{fig:gate_chn}
\end{figure}

\subsubsection{Spatial-wise selection gate module} 
	
The gate unit for spatial-wise selection outputs a 2-dimensional (2D) map $\bm{G}$ of size $(h_g,w_g,c_g)=(h,w,1)$. It will perform an element-wise product with the RoI feature maps $\bm{R}$ which is of size $h \times w \times c$ through a 1 by 1 convolution. As shown in Figure \ref{fig:gate_spatial}, through 1 by 1 convolution, the resulting 2D map has the same spatial resolution as the input feature maps. The 2D map is then passed through two fully connected (fc) layers and a Sigmoid function for normalization. The obtained 2D spatial mask $\bm{G}$ will be used to modulate the feature representation for every spatial location of the input feature. 
The feature values from all $C$ feature channels at spatial location $(i,j)$
will be modulated by the coefficient $\bm{G}(i,j,1)$. 
	

\subsubsection{Channel-wise selection gate module}
	
The gate model for channel-wise section generates a vector of size $(h_g,w_g,c_g) = (1,1,C)$ through depth-wise separable convolution \cite{MobileNetsV1}. 
As shown in Figure \ref{fig:gate_chn}, this vector is further passed through two fc layers and a Sigmoid function. The obtained $\bm{G}$ thereafter is used to perform a modulation with the convolutional features along the channel dimension. All the feature values within the $k$-th ($k\in [1,C]$) channel will be modulated by the $k$-th coefficient of $\bm{G}(1,1,k)$.

\section{Experiments}
\label{sec:experiment}

\subsection{Dataset and Experimental Setups}
\label{sec:settings}

 CityPersons \cite{CityPersons} is a recent pedestrian detection dataset built on top of the CityScapes dataset\cite{Cityscapes} which is for semantic segmentation. The dataset includes 5, 000 images captured in several cities of Germany. There are about 35, 000 persons with additional around 13, 000 ignored regions in total. Both bounding box annotation of all persons and annotation of visible person parts are provided. We conduct our experiments on CityPersons using the reasonable train/validation sets for training and testing, respectively. 


\noindent\textbf{Evaluation metrics}: Evaluations are measured using the log average missing rate (\textit{MR}) of false positive per image (FPPI) ranging from $10^{-2}$ to $10^{0}$ (\textit{MR}$_{-2}$). We evaluated 
four subsets with different ranges of pedestrian height ($hgt$) and  different visibility levels ($vis$) as follows: 

1)  All: $hgt\in[20, \inf]$ and $vis\in[0.2,\inf]$,

2) Small (Sm): $hgt\in[50, 75]$ and $vis\in[0.65,\inf]$,

3) Occlusion (Occ): $hgt\in[50, \inf]$ and $vis\in[0.2,0.65]$,

4) Reasonable (R): $hgt\in[50, \inf]$ and $vis\in[0.65,\inf]$.

\noindent \textbf{Network training and experimental setup}: 
The loss function contains a classification loss term and a regression loss term as in Faster-RCNN \cite{fastrcnn15}. Stochastic gradient descent with momentum is used for loss function optimization. A single image is processed at once, and for each image there are 256 randomly sampled anchors used to compute the loss of a mini-batch. The momentum $\lambda$ is set as 0.9 and and weight decay is set as $5\times 10^{-4}$.
Weights for the backbone network (i.e., $Conv1 \scriptsize{\sim} Conv5$ convolutional blocks) are initialized from the network pre-trained using the ImageNet dataset \cite{imagenet_dataset09}, while the other convolutional layers are initialized as a Gaussian distribution with mean 0 and standard deviation $0.01$.   We set the learning rate to $10^{-3}$ for the first 80k iterations and  $10^{-4}$ for the remaining 30k iterations. All experiments are performed on a single TITANX Pascal GPU. 

\subsection{Effectiveness of Squeeze Ratio}

\label{sec:sqeeze_ratio}
The squeeze ratio $r$ effects the network in terms of feature capacity and computational cost. To investigate the effects of squeeze ratio, we conduct experiments using features from multiple convolutional layers that have been squeeze by $r=1,2,4,8$ which will reduce the number of parameters in the following RoI-wise sub-networks by a factor of $1,2,4$ and 8, accordingly. The performances are compared in Table \ref{tab:sq_ratio}. We find that squeeze network can reduce the RoI-wise sub-network parameters without noticeable performance deduction. We use the reduction ratio $r = 2$ which is a good trade-off between performance and computational complexity. 

\begin{table}[t]
\center \begin{tabular}{c|c|c|c|c}
\hline
squeeze ratio  & All & Small & Occlusion  & Reasonable   \\ \hline
$r=1$     & 43.70      & 39.65 & 56.97   & 14.49     \\ \hline
$r=2$      & 43.02        & 42.02 & 55.60  & 14.35    \\ \hline
$r=4$      & 42.93       & 44.33 & 56.34   & 14.63   \\ \hline
$r=8$     & 44.52     & 39.98 & 58.06  & 14.85      \\ \hline
\end{tabular} 
\label{tab:sq_ratio}
\caption{Missing rate (MR\%) on Citypersons validation set using different squeeze ratios $r$.}
\end{table}

\subsection{Effectiveness of the Proposed Gate Models}
\label{sec:experiment_compare_gates}

\noindent \textbf{Baseline}: we use a modified version of the Faster-RCNN \cite{fasterRCNN2015} as our baseline detector. To generate pedestrian candidates, we use anchors of a single ratio of $\gamma=0.41$ with $9$ scales for the region proposal network. The baseline detector only adopts the $Conv5$ feature maps for feature representation. The limited feature resolution of $Conv5$ restraints the capability for detecting small pedestrians. We dilate the $Conv5$ features by a factor of two which enlarges the receptive field without increasing the filter size. 

For our ``spatial-wise gate" model and ``channel-wise gate", we use features extracted from the proposed gated multi-layer feature extraction network applying the two gate models, respectively. As can be seen from Table \ref{tab:gate_models_citypersons}, both the spatial-wise gate model and the channel-wise gate model make improvements upon the {Baseline} detector. These results demonstrate the effectiveness of our proposed gated multi-layer feature extraction. More specifically, the spatial-wise gate model achieves better performance on the ``\textit{Occlusion}" subset, while the channel-wise gate model achieves better performance on the ``\textit{Small}" subset.


\begin{table}[t]
\center
\begin{tabular}{c|c|c|c|c}
\hline 
Model  & All & Sm & Occ  & R   \\ \hline
FRCNN [{Baseline}] & 44.6   & 40.46  & 56.19  & 16.44\\ \hline
Adapted FRCNN \cite{CityPersons}  & - & - & - & 15.40\\ \hline
Repulsion Loss \cite{repulsion_loss2018}  & 44.45 & 42.63 & 56.85  & 13.22 \\ \hline
OR-CNN \cite{Occ-awareRCNN2018}  & 42.32 & 42.31 & 55.68 & \textbf{ 12.81 }\\ \hline
Spatial-wise gate     & \textbf{40.65} &  41.17 & \textbf{52.37}   & 13.64  \\ \hline
Channel-wise gate  & 41.76  & \textbf{37.62}  & 53.53  & 13.49  \\ \hline
\end{tabular}
\caption{Comparison of pedestrian detection performance (in terms of MR\%) of our proposed gate model with state-of-the-arts (in terms of MR\%) on the Citypersons dataset.}
\label{tab:gate_models_citypersons}
\end{table}

We also compare our proposed pedestrian detector with several state-of-the-art pedestrian detectors in Table \ref{tab:gate_models_citypersons}, including 
Adapted FRCNN	 \cite{CityPersons},  Repulsion Loss \cite{repulsion_loss2018}, and Occlusion-aware R-CNN (OR-RCNN) \cite{Occ-awareRCNN2018}. Repulsion Loss \cite{repulsion_loss2018} and OR-RCNN \cite{Occ-awareRCNN2018} are recent pedestrian detection methods proposed to address the occlusion problems.
For fare comparison, all the performance are evaluated on the original image size ($1024\times 2048 $) of the CityPersons validation dataset. We can observe that both of the two proposed gated models surpass the other approaches under the ``All",  ``Occlusion" and ``Small"  settings. The most notable improvements are on the ``Occlusion" and the ``Small" subsets. 
Our channel-wise gate model exceed the state-of-the-art method OR-RCNN \cite{repulsion_loss2018} by a large margin of  $4.7\%$ on the ''Small" subset, which highlights the effectiveness of our method for small-size pedestrian detection. When it comes to the ``Occlusion" subset where includes some severely occluded pedestrian, our spatial-wise gate model achieves the best $\textit{MR}_{-2}$ performance of $52.18\%$), surpassing the second best pedestrian detector by $2.5\%$. 

\section{Conclusions}
\label{sec:conclusion}

In this paper, we proposed a gated multi-layer convolutional feature extraction network for robust pedestrian detection. 
Convolutional features from backbone network are fed into the gated network to adaptively select discriminative feature representations for pedestrian candidate regions. A squeeze unit for feature dimension reduction is applied before the gated unit for RoI-wise feature manipulation. Two types of gate models that we proposed can manipulate the feature maps in channel-wise manner and in spatial-wise manner, respectively. 
Experiment on the CityPersons pedestrian dataset demonstrate the effectiveness of the proposed method for robust pedestrian detection. 

\bibliographystyle{IEEEbib}
\small \bibliography{references}

\end{document}